\documentclass{article}
\usepackage[preprint]{spconf}
\usepackage{cite}
\usepackage{amsmath,amssymb,amsfonts}
\usepackage{algorithmic}
\usepackage{graphicx}
\usepackage{textcomp}
\usepackage[dvipsnames]{xcolor}
\usepackage{todonotes}

\usepackage{hyperref}
\hypersetup{
    colorlinks=true,
    linkcolor=red,
    filecolor=magenta,      
    urlcolor=blue,
    citecolor=blue,
}
\usepackage{subcaption}
\graphicspath{{./figures/}}

\usepackage{pgf}
\usepackage{soul}
\usepackage{xspace}

\DeclareMathOperator*{\argmin}{arg\,min}

\newcommand\enc[1]{\text{Enc}\left(#1\right)}
\newcommand\dec[1]{\text{Dec}\left(#1\right)}
\newcommand\quant[1]{\text{Q}\left( #1 \right)}
\newcommand\dquant[1]{\text{DQ}\left( #1 \right)}
\newcommand\qround[1]{\text{Q}_{USQ}\left( #1 \right)}
\newcommand\dqround[1]{\text{DQ}_{USQ}\left( #1 \right)}

\newcommand\qapprox[1]{\text{Q}_{noise}\left( #1 \right)}

\newcommand\hypenc[1]{\text{Enc}_{hyper}\left(#1\right)}
\newcommand\hypdec[1]{\text{Dec}_{hyper}\left(#1\right)}

\title{
    \vspace{-25px}
    Optimizing Learned Image Compression on Scalar and Entropy-Constraint Quantization
    \thanks{
            \copyright 2024 IEEE.  Personal use of this material is permitted. Permission from IEEE must be obtained for all other uses, in any current or future media, including reprinting/republishing this material for advertising or promotional purposes, creating new collective works, for resale or redistribution to servers or lists, or reuse of any copyrighted component of this work in other works.
        }
    }

\name{
    \vspace{-20px}
}
\address{
    \textit{Florian Borzechowski$^{\dagger}$, Michael Schäfer$^{\dagger}$, Heiko Schwarz$^{\dagger\ddag}$,}\\
    \textit{Jonathan Pfaff$^{\dagger}$, Detlev Marpe$^{\dagger}$, Thomas Wiegand$^{\dagger\S}$}\\
    \\
    \textit{$^{\dagger}$Fraunhofer Institute for Telecommunications – Heinrich Hertz Institute, Berlin, Germany}\\
    \textit{$^{\ddag}$ Institute of Computer Science, Free University of Berlin, Germany}\\
    \textit{$^\S$ Department of Telecommunication Systems, Technical University of Berlin, Germany}\\
    firstname.lastname@hhi.fraunhofer.de
}

\begin{document}
\ninept
\maketitle

\begin{abstract}
    The continuous improvements on image compression with variational autoencoders have lead to learned codecs competitive with conventional approaches in terms of rate-distortion efficiency.
    Nonetheless, taking the quantization into account during the training process remains a problem, since it produces zero derivatives almost everywhere and needs to be replaced with a differentiable approximation which allows end-to-end optimization.
    Though there are different methods for approximating the quantization, none of them model the quantization noise correctly and thus, result in suboptimal networks.
    Hence, we propose an additional finetuning training step: After conventional end-to-end training, parts of the network are retrained on quantized latents obtained at the inference stage.
    For entropy-constraint quantizers like Trellis-Coded Quantization, the impact of the quantizer is particularly difficult to approximate by rounding or adding noise as the quantized latents are interdependently chosen through a trellis search based on both the entropy model and a distortion measure.
    We show that retraining on correctly quantized data consistently yields additional coding gain for both uniform scalar and especially for entropy-constraint quantization, without increasing inference complexity.
    For the Kodak test set, we obtain average savings between $1\%$ and $2\%$, and for the TecNick test set up to $2.2\%$ in terms of Bj{\o}ntegaard-Delta bitrate.
\end{abstract}


\section{Introduction}
Lossy image compression using neural networks has developed into a promising coding technology \cite{JPEGAI, lu2021transformerbased}.
The approach offers a competitive alternative to modern coding standards like VVC \cite{VVC} or HEVC \cite{sullivan2012overview} while its compression efficiency still continues improve.
Variational autoencoders (VAEs) \cite{TransformerBasedCoding, balle2017endtoend} are the most prominent instances of these networks.
They non-linearly transform the input image into latents which are then quantized and encoded using a jointly optimized entropy model and arithmetic coding \cite{arithmeticCoding}.
The decoder performs another non-linear transform to reconstruct an approximation of the input image.

The aforementioned entropy model uses another nonlinear network to estimate the probability distributions of the quantized latents which are then used at the arithmetic coding stage.
A well established approach for this estimation is applying a hypercoder network \cite{balle2018variational, MinnenBalleTodericiHyperprior}, which further compresses the latents into side information which is transmitted together with the latents.
The distributions of the latent variables are approximated as normal distributions and both encoder and decoder calculate estimates of the means and variances from the transmitted side information.
The arithmetic coder uses the probabilities obtained from these estimated distributions which vastly improves overall rate-distortion performance.
This basic approach is further improved by incorporating an autoregressive probability estimation or by transmitting additional side information to estimate the parameters of more complex distributions \cite{MinnenSinghCWAutoregressive, GaussianMixture}.
The models are trained using an end-to-end-approach, allowing the different parts of the neural network to be jointly optimized with respect to minimizing the rate-distortion cost \cite{RDOptimizedDeepImageCompression}.

However, taking the quantization into account during the end-to-end training process is nontrivial.
Any quantization function has zero gradients almost everywhere, impeding the backpropagation of the learning process with respect to the weights of the encoding network and therefore disallowing end-to-end training.
Hence, quantization needs to be replaced with an approximation of its impact on the latents.
Previous papers use noise perturbation \cite{balle2018variational, MinnenBalleTodericiHyperprior, EtEIntroduction}, straight-through estimation \cite{theis2017lossy, Liu_2022_CVPR} and similar other approximations \cite{MinnenSinghCWAutoregressive, Pan_2021_CVPR, QuantComparisons}.
The findings of \cite{QuantComparisons} indicate that there is no single optimal approximation, but different approximations work better for different network architectures.
However, those works mostly relate to uniform scalar quantization (USQ) while hard to approximate entropy-constraint vector quantizers such as trellis coded quantization (TCQ) \cite{TCQ,OtherTCQ} are yet to be explored by these methods.

This paper aims to circumvent the approximation step by using a pre-trained model as an anchor and finetuning parts of the network on quantized latent data aquired through proper application of the quantization processes.
This is similar to the two-step training suggested in \cite{guo2021soft}, which uses rounding to integers for decoder refinement.
However, we approach this explicitly to allow more sophisticated entropy-constraint quantization mechanisms than USQ to be finetuned in the retraining process, which to the best of our knowledge has not been explored before.

In particular, we consider TCQ, which is more complex than simple rounding because it implements a low-complexity vector quantization \cite{VectorQuant} in which quantization decisions are made based on previous samples, and their assumed impact on the bitrate and distortion.
The quantization therefore depends on past quantization decisions, a distortion metric, and the entropy model, i.e. the hypercoder network.
For selecting a suitable distortion measure, \cite{TCQ} have demonstrated that instead of using sample distortion, through which updates of the decoder network would have an inevitable impact on the quantization decisions, it suffices to consider the distortion between original and reconstructed latents.
They have shown that using a simple cost criterion with an empirically chosen balance between bitrate and the latents' quantization error yields coding gains of more than $2\%$ over uniform scalar quantization.
Furthermore, as TCQ uses multiple scalar quantizers whose reconstruction levels are not necessarily uniform, conventional uniform noise perturbation might not optimally reflect the distribution of the quantized latents anymore.
These complexities of the quantization decisions make TCQ considerably harder to approximate in a differential manner, and let TCQ therefore pose as a potent example for effective quantization finetuning of complicated quantization schemes.

For our experiments, adequately pre-trained VAEs are used as a basis and applicable parts of the networks are then further retrained.
We optimize the decoder network with respect to the distortion by applying either TCQ or USQ to the latents as determined by the pre-trained encoder from the inference stage.
In the case of USQ, the entropy model is additionally retrained with respect to minimizing the expected bitrate.
Contrasting \cite{guo2021soft}, which likewise deploys multiple training stages with soft and hard quantization, this approach allows to consider non-uniform, entropy-constraint quantizers.
We show that this retraining step on encoder-determined quantized latents leads to improvements for both USQ and TCQ.
In particular, the relative Bj{\o}ntegaard-Delta bitrate (BD-rate) improvement \cite{bjontegaard2001calculation} in the case of our four-state TCQ implementation is consistently higher than in the USQ case.
This indicates that rate-distortion optimized quantization is indeed more difficult to approximate by smooth approximations as stated in \cite{TCQ}, but still able to be effectively optimized with our finetuning step.
As our improvements only adjust existing weights, the gains come without increasing the runtime complexity at inference.

\begin{figure*}[t]
    \centering
    \includegraphics[width=0.8\textwidth]{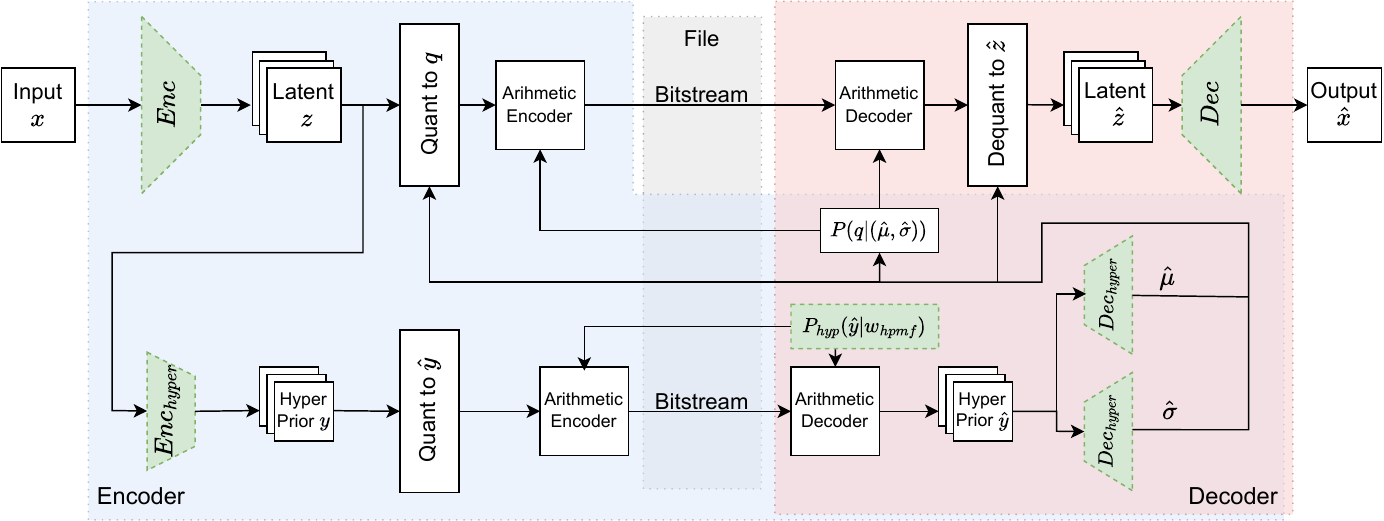}
    \caption{Comprehensive Flowchart of the used auto encoder. Green, dashed elements employ trainable weights. The quantization blocks use $\mu$ either for mean shifting as in \eqref{eq:usq}, or for the rate-constraint in TCQ.}
    \label{fig:architecture}
\end{figure*}

The remainder of the paper is organized as follows.
Section \ref{section:description} describes the architecture of the investigated VAE network, the design of the employed quantizers, and how these are substituted in the initial training step.
Section \ref{section:experiments} then presents and discusses the results of our retraining experiments.
The paper concludes with our main findings in Section \ref{section:conclusion}.

\section{Network Architecture}\label{section:description}

\subsection{General autoencoder description}

The network used in this paper is based on the autoencoder design of \cite{RDOptimizedDeepImageCompression, TCQ} and follows the basic architecture of VAEs, where an input image $x$ is transformed by an encoder network to the latent representation $z \in \mathbb{R}^N$, where $N$ describes the total number of latent coefficients; see Figure \ref{fig:architecture}.
A quantization function is then applied, converting the latents to quantization indices $q \in \mathbb{Z}^N$, usually symmetric around the estimated means $\hat{\mu}$. At the decoder side, the quantization indices are dequantized into the reconstructed latent $\hat{z} \in \mathbb{R}^N$, which can then be transformed by a decoder network, yielding the approximation $\hat{x}$ of the input image:
\begin{equation}\label{eq:coder}
    \begin{aligned}
        z       & = \enc{x; w_{enc}},       \\
        q       & = \quant{z, \hat{\mu}},   \\
        \hat{z} & = \dquant{q, \hat{\mu}},  \\
        \hat{x} & = \dec{\hat{z}; w_{dec}}.
    \end{aligned}
\end{equation}
As with common VAE implementations, our encoder and decoder networks are based on convolutional layers and GDN nonlinearities \cite{balle2016density}, the trainable weights of which being represented by $w_{enc}$ and $w_{dec}$. Additionally, we utilize octave convolutions in three different resolution levels similar to \cite{OctaveConvolutions} to further reduce spatial redundancies in the latent representation.

For transmitting the quantization indices $q$ with an arithmetic coder, a hypercoder network is applied to the latent to estimate the probability distributions for the arithmetic coding process. The hypernetwork creates the hyperprior $y$ from the unquantized latent $z$ and quantizes the hyperprior to $\hat{y}$ with
\begin{equation}\label{eq:hyperencoder}
    \begin{aligned}
        y       & = \hypenc{z; w_{hypenc}}, \\
        \hat{y} & = \lfloor y \rceil.
    \end{aligned}
\end{equation}
The hyperdecoder network uses the quantized hyperprior $\hat{y}$ to generate estimates $\hat{\mu}$ and $\hat{\sigma}$ of the mean and the standard deviation for each coefficient of $z$:
\begin{equation}\label{eq:hyperdecoder}
    \begin{aligned}
        \hat{\mu}    & = \hypdec{\hat{y}; w^{mean}_{hypdec}}, \\
        \hat{\sigma} & = \hypdec{\hat{y}; w^{dev}_{hypdec}}.
    \end{aligned}
\end{equation}

It is assumed that the unquantized latents are normal distributed. Then, for each entry $q_i$ of the quantization indices $q$, the probability mass function (pmf) used for arithmetic coding is obtained by quantizing the Gaussian distribution $\mathcal{N}(\hat{\mu}, \hat{\sigma}^2)$ according to:
\begin{equation}\label{eq:inference-latent-probs}
    P(q_i | (\hat{\mu}_i, \hat{\sigma}_i)) = \int^{b}_{a} \mathcal{N}(t; \hat{\mu}_i, \hat{\sigma}_i^2) dt,
\end{equation}
where the integration limits are:
\begin{equation}\label{eq:inference-prob-limits}
    \begin{aligned}
        a & = \dquant{q_i, \hat{\mu}_i} - \Delta/2, \\
        b & = \dquant{q_i, \hat{\mu}_i} + \Delta/2.
    \end{aligned}
\end{equation}
Here, $\Delta$ denotes the quantization step size.
The hyperprior $\hat{y}$ is likewise encoded by the arithmetic coder, its distribution is estimated by a learned, static probability distribution $P_{hyp}(\cdot | w_{hpmf})$ with the learned weights $w_{hpmf}$, and a step size of $\Delta = 1$, as demonstrated in \cite{balle2018variational}.
All weight variables used in the network will be collectively referred to as:
\begin{equation}
    w =
    \left[w_{enc}, w_{dec}, w_{hypenc},  w_{hpmf},w^{mean}_{hypdec}, w^{dev}_{hypdec}\right].
\end{equation}

\subsection{Uniform-scalar and entropy-constraint quantization}\label{section:quantfunction}

In common VAE designs, USQ is chosen for quantization at inference. The latents are quantized by dividing each value by the quantization step size $\Delta$ and rounding the result to the next integer; the dequantization process then maps the quantization indices to the uniformly spaced reconstruction levels:
\begin{equation}\label{eq:usq}
    \begin{aligned}
        q       & = \qround{z, \hat{\mu}} = \lfloor (z - \hat{\mu}) / \Delta \rceil, \\
        \hat{z} & = \dqround{q, \hat{\mu}} = q \cdot \Delta + \hat{\mu}.
    \end{aligned}
\end{equation}
Note that during inference, we subtract the estimated mean $\hat{\mu}$ before the quantization, and add it back after the dequantization.
Particularly for low-rate operation points, at which many values are quantized to zero, this typically decreases the average quantization error.

Some approaches use more advanced quantizers, such as TCQ \cite{TCQ,OtherTCQ} which represents a complexity-constrained vector quantizer \cite{VectorQuant}.
We follow the TCQ design of \cite{TCQ, QuantInVVC} where from a decoder perspective, TCQ employs two distinct scalar quantizers and a state transition table.
Given a fixed scan order of the latents, the quantizer is selected based on the parity of the preceding quantization index.
In our case, the first quantizer has the reconstruction levels $\{0, -2\Delta, 2\Delta, -4\Delta, 4\Delta, ...\}$ and is hence called even, while the second one has levels $\{0, -\Delta, \Delta, -3\Delta, 3\Delta, ... \}$ and is referred to as odd.
Note that both quantizers contain the zero level which yields improved low-rate coding efficiency.
Given a measure of the rate-distortion cost per quantization index, the possible encoder decisions can be represented by a trellis structure. The Viterbi algorithm then solves the encoding problem of finding the rate-distortion-optimal path through the trellis \cite{Viterbi}.
The rate is calculated with the estimates $\hat{\mu}$ and $\hat{\sigma}$, which are used to compute probability tables for each quantizer with their distinct quantization intervals.
The reconstruction error is estimated by using the latent quantization error $||\hat{z}-z||^2$.

\subsection{Initial training with smooth approximation of the quantizer}\label{section:training}

When initially training the entire VAE network, the quantization function is replaced with a differentiable approximation to allow for useful gradients to be calculated for the encoder network.
For VAEs meant to use USQ during inference, a usual replacement \cite{EtEIntroduction} is adding uniform white noise onto the latents, simulating the average perturbation caused by true quantization:
\begin{equation}\label{eq:qapprox}
    \tilde{z} = \qapprox{z} = z + \mathcal{U}\left(-\frac{\Delta}{2}, \frac{\Delta}{2}\right).
\end{equation}
This process is based on the assumption that the perturbations caused by real quantization are uniformly distributed, though it can be argued that this representation is too simple and not realistic enough to allow for optimal training \cite{QuantComparisons,guo2021soft}.

Approximating the behaviour of the latent perturbation in TCQ quantization proves to be more difficult.
Previous works \cite{TCQ} attempt to reflect the complex decision process at the training stage by perturbing the latent with uniformly distributed white noise, switching between two different perturbations $\bar{z}_i^0 := z+2\Delta\ \mathcal{U}(-0.5, 0.5)$ and $\bar{z}^1 := \bar{z}^0 - \text{sgn}(\bar{z}^0) \Delta$.
The minimizer of the error $\bar{z}:=\operatorname{arg}\ \operatorname{min}_{t\in\{\bar{z}^0,\bar{z}^1\}} |z-t|$ is then selected for each entry of the latents and used as input of the decoder network.

In all cases, the quantization of the hyperprior with its fixed quantization step size of $\Delta = 1$ is approximated by
\begin{equation}\label{eq:hyperapprox}
    \tilde{y} =  y + \mathcal{U}(-0.5, 0.5).
\end{equation}
The estimated mean and variance parameters calculated with the approximately quantized hyperprior $\tilde{y}$ are referred to as $\tilde{\mu}$ and $\tilde{\sigma}$, respectively.

Using rate-distortion loss as a training objective, the autoencoder network is jointly optimized as follows:
\begin{equation}\label{eq:loss}
    w^* = \argmin_{w} \left( D + \lambda \cdot R \right).
\end{equation}
Here, $D$ denotes the distortion
\begin{equation}
    D = \sum_k^K (x_k - \dec{\tilde{z}; w_{dec}}_k)^2,
\end{equation}
where $k$ is an flattened index of the samples of the input image.
The bitrate $R$ is approximated as
\begin{equation}
    R = \sum_{i}^N -\log_2{P(\tilde{z}_i | \tilde{y}_i)}        \\
    + \sum_{j}^M -\log_2{P_{hyp}(\tilde{y}_j | w_{hpmf})},
\end{equation}
with $i$ and $j$ sequentially flat-indexing the multidimensional latents of size $N$ and hyperpriors of size $M$, respectively.
For the TCQ case, we simply replace $\tilde{z}$ by $\bar{z}$ at the training stage.
The Lagrange multiplier $\lambda$ is used to prioritize either the bitrate or the distortion, effectively shifting the trained model to higher or lower operation points of the rate-distortion curve.

The estimation of the bitrate is similar to the calculation of the symbol probabilities during inference as described in \eqref{eq:inference-latent-probs}.
However, instead of using quantization indices as in \eqref{eq:inference-prob-limits}, the probability distribution is calculated with the limits of $a = \tilde{z} - \Delta/2$ and $b = \tilde{z} + \Delta/2$.
It is implicitly assumed that the arithmetic coder achieves the entropy limit given by the estimated probability mass functions.

\section{Training experiments on quantized latents determined by the encoder}\label{section:experiments}

The experimental VAEs were trained on a subset of the Imagenet dataset \cite{ImageNet}, with cropped $256\times256$ luma blocks.
Inference results were calculated on fully sized Kodak images \cite{Kodak} of resolution $768\times512$, and TecNick images \cite{tecnick} with a resolution of $1200 \times 1200$.
The experiments were implemented, and optimization was performed, using the Tensorflow Deep Learning package \cite{Tensorflow}.
One epoch consists of $250$ batches with $8$ (for the USQ models) and $4$ (for the TCQ models) examples per batch.
Whenever loss saturation was achieved, the learning rate  $l_{i+1}=\alpha l_i$ was decreased, starting from $l_0 = 10^{-6}\alpha$ with a decay factor of $\alpha=(1.131)^{-1}$.
On machines with four Ampere A100 GPUs and 40GB of RAM, the retraining for the USQ models took four hours at most per network.
The retraining steps for the TCQ networks lasted eight hours due to the higher IO load of the generated TCQ dataset.

Five VAEs for different operation points on the rate-distortion curve were pre-trained by setting the Lagrange parameter  $\lambda\in\{128,256,512,1024,2048\}$.
For our experiments, two sets of VAEs were employed.
The first set is derived from \cite{RDOptimizedDeepImageCompression}; its networks were optimized on uniformly-noised latents for approximating USQ.
The second is derived from the \cite{TCQ} set and was specifically trained for TCQ by switching between approximations $\bar{z}^0$ and $\bar{z}^1$ as described in \ref{section:training}.
PSNR-curves visualizing all results are provided in Figure \ref{fig:PSNRCurves}.
BD-rates of the experiments are comprehensively listed for Kodak, and averages for the TecNick set are given in Table \ref{tab:tecnickbd}.

\subsection{Retraining of the decoder}\label{section:decoderretrain}

\begin{table}
    \centering
    \begin{tabular}{|r|rr||r|rr|}
        \hline
        Img & high    & low     & Img           & high             & low              \\ \hline
        1   & -0.62\% & -0.95\% & 13            & -0.52\%          & -0.87\%          \\ \hline
        2   & -1.22\% & -1.51\% & 14            & -0.75\%          & -1.11\%          \\ \hline
        3   & -1.13\% & -1.38\% & 15            & -0.92\%          & -1.13\%          \\ \hline
        4   & -0.95\% & -1.34\% & 16            & -0.80\%          & -1.14\%          \\ \hline
        5   & -0.85\% & -1.27\% & 17            & -1.03\%          & -1.36\%          \\ \hline
        6   & -0.71\% & -1.08\% & 18            & -0.73\%          & -1.09\%          \\ \hline
        7   & -1.09\% & -1.51\% & 19            & -0.84\%          & -1.29\%          \\ \hline
        8   & -0.81\% & -1.08\% & 20            & -1.27\%          & -1.91\%          \\ \hline
        9   & -0.98\% & -1.29\% & 21            & -0.73\%          & -1.12\%          \\ \hline
        10  & -1.06\% & -1.35\% & 22            & -0.70\%          & -1.03\%          \\ \hline
        11  & -0.84\% & -1.27\% & 23            & -1.05\%          & -1.40\%          \\ \hline
        12  & -1.14\% & -1.54\% & 24            & -0.64\%          & -1.03\%          \\ \hline
        \hline
            &         &         & \textbf{Avg.} & \textbf{-0.87\%} & \textbf{-1.24\%} \\ \hline
    \end{tabular}
    \caption{BD-rates of Kodak images encoded with the retrained USQ decoder model compared to the USQ anchor model.}
    \label{tab:USQBase-vs-USQDec}
\end{table}

The first experiment aimed to optimize only the decoder network $w_{dec}$ of the VAE using actual quantized data. For this, the models trained for USQ were retrained by replacing the quantization approximation function \eqref{eq:qapprox} with the true quantization function \eqref{eq:usq}, including the mean shift.
The training was then conducted with only the weights of the decoder $w_{dec}$ being allowed to be adjusted, with the encoder and the entire hypercoder being effectively frozen.
This excludes the rate term from the training process and simplifies the training objective to:
\begin{equation}
    w_{dec}^{*} = \argmin_{w_{dec}; w } D.
\end{equation}

The BD-rates achieved by this retraining are listed in Table \ref{tab:USQBase-vs-USQDec}.
The BD-rate is a measure that compares a PSNR-rate curve (given by a set of data points) against a reference PSNR-rate curve and specifies the average relative rate difference to the reference curve for the same PSNR, where negative values indicated bitrate savings \cite{bjontegaard2001calculation}.
Retraining the decoder leads to a BD-rate of $-0.87\%$ for high bitrates and $-1.24\%$ for low bitrates, meaning that the retraining improved the model's performance. The models trained for low bitrates generally benefit more from the retraining with stronger BD-rate gain. A possible explanation of this behavior is that the application of larger quantization step sizes in the quantization approximation \eqref{eq:qapprox} conform less to the assumption of uniformly distributed quantization noise than smaller quantization step sizes, leading to a larger approximation error to be corrected through retraining.
It is notable that the improvements are consistent over the entire testing dataset, with every operation point being improved by the retraining, and no decreases in quality occurring.

Since only the decoder is retrained, the bitstreams of the test images are the same as for the pre-trained model, and only the reconstruction error decreases. On average, the improvement of the PSNR in the Kodak dataset between the anchor model and the model with the retrained decoder is $0.072$ dB.

\begin{table}
    \centering
    \begin{tabular}{|r|rr||r|rr|}
        \hline
        Img & high    & low     & Img           & high             & low              \\ \hline
        1   & -1.84\% & -1.63\% & 13            & -1.89\%          & -1.49\%          \\ \hline
        2   & -1.78\% & -1.71\% & 14            & -1.74\%          & -1.77\%          \\ \hline
        3   & -2.15\% & -2.11\% & 15            & -1.83\%          & -1.86\%          \\ \hline
        4   & -2.04\% & -2.00\% & 16            & -2.07\%          & -1.83\%          \\ \hline
        5   & -2.25\% & -2.27\% & 17            & -2.07\%          & -2.10\%          \\ \hline
        6   & -2.08\% & -1.92\% & 18            & -1.95\%          & -1.66\%          \\ \hline
        7   & -2.26\% & -2.31\% & 19            & -1.80\%          & -1.76\%          \\ \hline
        8   & -1.72\% & -1.69\% & 20            & -1.83\%          & -2.19\%          \\ \hline
        9   & -1.96\% & -1.98\% & 21            & -2.01\%          & -1.83\%          \\ \hline
        10  & -2.10\% & -2.12\% & 22            & -1.72\%          & -1.62\%          \\ \hline
        11  & -1.89\% & -1.86\% & 23            & -2.67\%          & -2.44\%          \\ \hline
        12  & -1.89\% & -1.88\% & 24            & -1.64\%          & -1.42\%          \\ \hline
        \hline
            &         &         & \textbf{Avg.} & \textbf{-1.97\%} & \textbf{-1.90\%} \\ \hline
    \end{tabular}
    \caption{BD-rates of Kodak images encoded with the retrained TCQ decoder model compared to the TCQ anchor model.}
    \label{tab:TCQBase-vs-TCQDec}
\end{table}

\subsection{Decoder retraining for TCQ}\label{section:TCQretraining}

Because of TCQ being a more advanced quantization scheme than USQ, it is highly interesting to do a retraining on true data. Only thusly the complex interdependencies between samples can be passed through to the decoder, which could otherwise not be provided by per-sample approximations. As such, the same experiment was conducted for the TCQ-optimized VAEs.
The used TCQ design is based on the tradeoff between rate and latent space distortion as in \cite{TCQ} which allows for updates to the decoder network without impact on how the distortion is measured at the quantization stage.
Replacing the approximation function with a true TCQ process is nontrivial, so TCQ was applied offline to the entire dataset, pre-generating a comprehensive pool of TCQ reconstructed features $\hat{z}$.
This pre-generated dataset was then loaded during the training process, allowing the decoder to train on TCQ-quantized data.
However, since the selection of the quantization indices heavily depends on the entropy model, any fixed dataset cannot account for updates of the hypercoder network.
This impedes a proper retraining of the hypercoder network.

The experimental results of the retrained decoder for TCQ models in comparison to the original anchor models are listed in Table \ref{tab:TCQBase-vs-TCQDec}.
Similar to the previous experiment, the improvement is consistent over all rate points and test images, showing larger improvements than USQ, with an average BD-rate of $-1.97\%$ for high bitrates, and $-1.90\%$ for low bitrates over the Kodak image set.
Testing on the higher resolution TecNick set produced BD-rates of $-1.98\%$ and $-2.29\%$ as listed in Table \ref{tab:tecnickbd}.
As with USQ, retraining the TCQ decoder only leads to reconstruction improvements. The average PSNR increase of all rate points across the Kodak dataset is $0.136$ dB, and $0.153$ dB over the TecNick set, with no model decreasing in quality at any point.

\subsection{Retraining of the hypercoder network}

\begin{table}
    \centering
    \begin{tabular}{|r|rr||r|rr|}
        \hline
        Img & high    & low     & Img           & high             & low              \\ \hline
        1   & -0.81\% & -1.37\% & 13            & -0.65\%          & -1.09\%          \\ \hline
        2   & -1.59\% & -2.44\% & 14            & -0.89\%          & -1.42\%          \\ \hline
        3   & -1.47\% & -2.34\% & 15            & -1.25\%          & -1.78\%          \\ \hline
        4   & -1.16\% & -1.85\% & 16            & -0.82\%          & -1.36\%          \\ \hline
        5   & -1.01\% & -1.59\% & 17            & -1.26\%          & -1.94\%          \\ \hline
        6   & -0.89\% & -1.41\% & 18            & -0.92\%          & -1.33\%          \\ \hline
        7   & -1.27\% & -1.95\% & 19            & -1.07\%          & -1.67\%          \\ \hline
        8   & -1.03\% & -1.51\% & 20            & -2.83\%          & -3.89\%          \\ \hline
        9   & -1.31\% & -1.96\% & 21            & -0.89\%          & -1.46\%          \\ \hline
        10  & -1.39\% & -1.98\% & 22            & -0.82\%          & -1.32\%          \\ \hline
        11  & -1.02\% & -1.67\% & 23            & -1.18\%          & -1.69\%          \\ \hline
        12  & -1.39\% & -2.11\% & 24            & -0.86\%          & -1.43\%          \\ \hline
        \hline
            &         &         & \textbf{Avg.} & \textbf{-1.14\%} & \textbf{-1.73\%} \\ \hline
    \end{tabular}
    \caption{BD-rates of Kodak images encoded with the USQ model with a retrained hypercoder network and decoder, compared to the USQ anchor model.}
    \label{tab:USQBase-vs-USQHypCoderDec}
\end{table}

\begin{table}[t]
    \centering
    \begin{tabular}{|l|rr|}
        \hline
        Model            & High    & Low     \\ \hline
        USQ Decoder      & -1.06\% & -1.56\% \\
        USQ HypCoder+Dec & -1.36\% & -1.99\% \\
        TCQ Decoder      & -1.98\% & -2.29\% \\ \hline
    \end{tabular}
    \caption{Average BD-rates between retrained models and their respective anchors for the TecNick test dataset.}
    \label{tab:tecnickbd}
\end{table}

\begin{figure*}[t]
    \begin{subfigure}[b]{0.5\textwidth}
        \resizebox{1\textwidth}{!}{\input{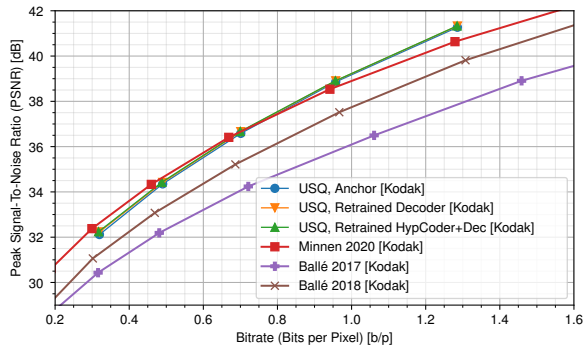}}
        \caption{USQ model performance on the Kodak test set.}
        \label{fig:USQKodak}
    \end{subfigure}
    \begin{subfigure}[b]{0.5\textwidth}
        \resizebox{1\textwidth}{!}{\input{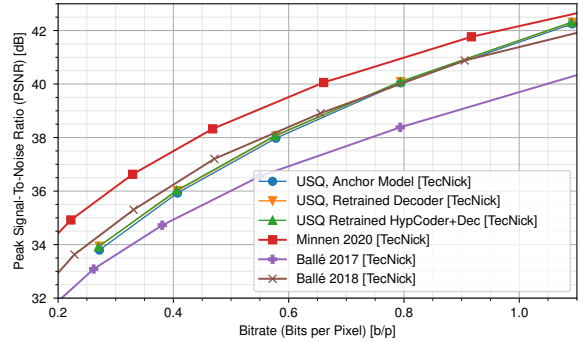}}
        \caption{USQ model performance on the TecNick test set.}
        \label{fig:USQTecNick}
    \end{subfigure}
    \begin{subfigure}[b]{0.5\textwidth}
        \resizebox{1\textwidth}{!}{\input{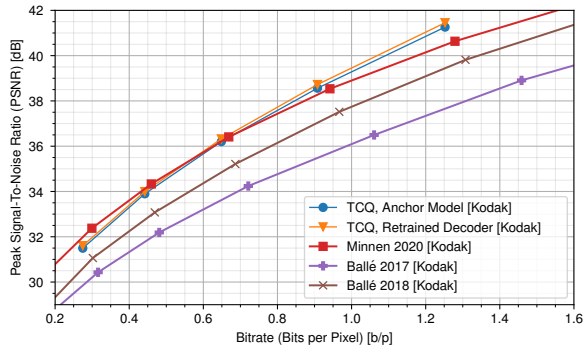}}
        \caption{TCQ model performance on the Kodak test set.}
        \label{fig:TCQKodak}
    \end{subfigure}
    \begin{subfigure}[b]{0.5\textwidth}
        \resizebox{1\textwidth}{!}{\input{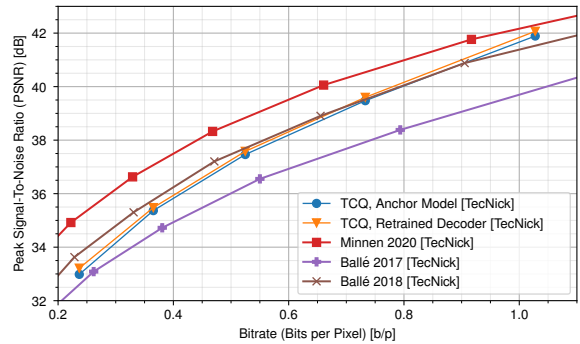}}
        \caption{TCQ model performance on the TecNick test set.}
        \label{fig:TCQTecNick}
    \end{subfigure}
    \caption{Average PSNR-curves of our models in comparison with Ballé17 \cite{balle2017endtoend}, Ballé18 \cite{balle2018variational}, and Minnen20 \cite{MinnenSinghCWAutoregressive} on Kodak and TecNick datasets.}
    \label{fig:PSNRCurves}
\end{figure*}

The decoder is not the only part of the VAE affected by approximating quantization functions.
The hypercoder network is being optimized to provide probability distribution estimates for the perturbed quantization indices $\tilde{z}$, which may be distributed differently to the truly quantized latents.
Therefore, a third experiment was set up, which retrained the entire hypercoder system as described in \eqref{eq:hyperencoder} and \eqref{eq:hyperdecoder} together with the decoder as in the previous experiment.
To allow for the hyperencoder to be trained as well, the quantization of the hyperprior $y$ was again replaced with the approximation \eqref{eq:hyperapprox}, and only the latent is truly quantized with USQ.
To include the hypercoder in the training process, the loss term for this retraining includes both the distortion and the rate, following \eqref{eq:loss}. This configuration would allow $\tilde{\mu}$ and $\tilde{\sigma}$ to adjust to the real quantization indices instead of their continuous approximations through the rate term. Out of all weights $w$, the weight parameters $w_{hypenc}$, $w_{hpmf}$, $w^{mean}_{hypdec}$, $w^{dev}_{hypdec}$, and $w_{dec}$ were optimized.
It should be noted that $\tilde{\mu}$ is unable to gather useful gradients for the optimization process in relation to the distortion.
Therefore, the gradient calculation for $\tilde{\mu}$ was intentionally zeroed out in the quantization functions \eqref{eq:usq}, allowing the mean to only be optimized with respect to the rate.

The BD-rate results of this experiment are listed in Table \ref{tab:USQBase-vs-USQHypCoderDec}. Retraining both the hypercoder on rate and the decoder on distortion leads to a generally favorable BD-rate in comparison to the anchor model of $-1.14\%$ for high bitrates, and $-1.73\%$ for low bitrates. This also constitutes an improvement over the pure decoder retrain of \ref{section:decoderretrain}, against which a BD-rate of $-0.27\%$ and $-0.49\%$ was measured.
This improvement over the first experiment is mostly due to ability of this configuration to jointly optimize rate and distortion. Similar to the distortion measure, the bitrate consistently decreased for every data point in almost every experiment (except for tested Kodak images 20, 21, and 22, which experienced slightly higher bitrates at high rate operation points). The average decrease in bitrate is $0.0015$ bits per sample.

\subsection{Discussion}

For the Kodak test set, our model performs similarly to Minnen 2020 \cite{MinnenSinghCWAutoregressive} as visible in figures \ref{fig:USQKodak} and \ref{fig:USQTecNick}, with better high rate performance and slightly decreased performance for lower bitrates, possibly because of octave convolutions allowing for a larger amount of high frequency detail being saved in the latents.
Further note that our entropy model does not employ an auto-regressive component similar to \cite{balle2017endtoend} and thus, has less degrees of freedom for estimating the latent probabilities than \cite{MinnenSinghCWAutoregressive}.
In the higher resolution TecNick test set, our model sees a general decrease in relation to Minnen 2020, performing similar to Ballé 2018 \cite{balle2018variational} instead.
The performance loss is possibly caused by the training set not containing enough higher-resolution data.
Regardless of the investigated test sets however, the improvements of the retraining step on quantized data from the encoder remain within the same magnitude over all operation points.
The consistency of the improvements over all operational points of the RD-curve even with different image sizes differs from the experimental quantization approximations of \cite{MinnenSinghCWAutoregressive}, where hard rounded latents were used for some synthesis transforms, leading to improvements mostly in the low bitrate range.

The retraining results even just for the decoder of the TCQ model are consistently higher than both retrainings for USQ. These interesting gains demonstrate that complex quantization schemes have potential to be further improved without increasing complexity through our retraining. We have shown that the decoder is able to process complicated interdependencies, which suggests similarly positive results for other quantization schemes as well.

\begin{table}[t]
    \centering
    \begin{tabular}{|lr|rrr|}
        \hline
                                        & \multicolumn{1}{l|}{}       & \multicolumn{1}{l}{USQ}          & \multicolumn{1}{l}{USQ}      & \multicolumn{1}{l|}{TCQ} \\
        \multicolumn{2}{|r|}{$\lambda$} & \multicolumn{1}{l}{Decoder} & \multicolumn{1}{l}{HypCoder+Dec} & \multicolumn{1}{l|}{Decoder}                            \\ \hline
                                        & 128                         & 0.118 dB                         & 0.123 dB                     & 0.112 dB                 \\
                                        & 256                         & 0.065 dB                         & 0.073 dB                     & 0.097 dB                 \\
        Kodak                           & 512                         & 0.077 dB                         & 0.084 dB                     & 0.119 dB                 \\
                                        & 1024                        & 0.045 dB                         & 0.047 dB                     & 0.162 dB                 \\
                                        & 2048                        & 0.055 dB                         & 0.059 dB                     & 0.192 dB                 \\ \hline \hline
                                        & 128                         & 0.147 dB                         & 0.127 dB                     & 0.231 dB                 \\
                                        & 256                         & 0.087 dB                         & 0.099 dB                     & 0.112 dB                 \\
        TecNick                         & 512                         & 0.087 dB                         & 0.100 dB                     & 0.118 dB                 \\
                                        & 1024                        & 0.044 dB                         & 0.047 dB                     & 0.125 dB                 \\
                                        & 2048                        & 0.058 dB                         & 0.063 dB                     & 0.177 dB                 \\ \hline
    \end{tabular}
    \caption{Absolute average PSNR differences between the USQ/TCQ experiments and their respective anchor models.}
    \label{tab:PSNRcomparisons}
\end{table}

\subsection{Outlook}\label{section:outlook}

Since TCQ uses the estimated normal distribution provided by the hypercoder to generate the quantized latents, optimizing the TCQ hypernetwork and decoder together would require computing $\hat{z}$ on-the-fly.
As the retrained hypercoder changes the trellis costs in the Viterbi algorithm at inference, it would also lead to a different quantization pattern which is not reflected by the static, offline latent dataset.
Thus, the current approach potentially leads to different distributions between the dequantized latents of the training data and inference data.
This dissonance remains to be remedied in future experiments.

However, the general approach of re-optimizing the decoder with respect to sample distortion produces positive results.
It leads to consistent PSNR increases as listed in Table \ref{tab:PSNRcomparisons}.
In particular, we measured larger improvements for the TCQ decoder retraining in comparison to the USQ case.
Note that TCQ selects quantization indices based on a suitable rate-distortion criterion, which possibly depend on the parameters of the network itself.
Thus, any quantization approximation, whether through noisy perturbation or hard or soft rounding, hardly provides accurate training data for the decoder.
Vice versa, the decoder appears to additionally benefit from using encoder-generated data in the case of TCQ.

Based on our observations, we predict that replacing soft approximations by quantized training data from the encoder lends itself to other complex rate-distortion-optimized quantization schemes, or even other architectures. As long as some quantization approximation happens, injecting truly quantized data into the training process should lead to successful finetuning.
Even in the case of uniform scalar quantization, positive results for rate-distortion optimized encoding have been achieved in \cite{RDOptimizedDeepImageCompression, Brand_2022_CVPR}.
This opens the possibility to extend the approach from \cite{guo2021soft} as we did with TCQ; After conventional end-to-end optimization using noisy or soft perturbation, rate-distortion-optimization on the encoded latents can be performed and used as training data for retraining the decoder.
Multiple repetitions of generating test data by applying true quantization with the updated decoder can be performed until convergence is achieved.

\section{Conclusion}\label{section:conclusion}

We show that incorporating a retraining of the decoder network into the training process of learned image compression networks further increases the reconstruction quality without increasing inference runtime.
Especially if the quantization function is more complicated and harder to approximate with a gradient-friendly function, optimizing the decoder on quantized latents from the encoder leads to better coding efficiency.
The reason for this improvement is the more accurate nature of the training data in comparison to conventional approximations with uniform noise or soft quantizers, which fail to capture the involved complexities of entropy-constraint samples.
The effect was verified in the case of uniform scalar quantization, leading to average PSNR gains above $0.1$ dB, and for a specific implementation of TCQ with PSNR increases of up to $0.23$ dB.
In terms of BD-rate, the retraining effect for TCQ accounts for an additional coding gain of about $2\%$ on luma-only versions of the Kodak and Tecnick dataset.

\bibliographystyle{./IEEEbib.bst}
\bibliography{./references/IEEEabrv, ./references/references.bib}
\end{document}